\documentclass[journal=jacsat,manuscript=article]{achemso}

\usepackage{chemformula} % Formula subscripts using \ch{}
\usepackage[T1]{fontenc} % Use modern font encodings
\usepackage{lineno}
\usepackage{times}
\usepackage{epsfig}
\usepackage{subfigure}
\usepackage{booktabs}
\usepackage{graphicx}
\usepackage{amsmath,mathrsfs}
\usepackage{amssymb}
\usepackage{bm}
\author{Bo Zhang}
\affiliation{Department of Automation, Tsinghua University, Beijing 100084, China}
\author{Yuqi Han}
\affiliation{Department of Automation, Tsinghua University, Beijing 100084, China}
\author{Jinli Suo}
\affiliation{Department of Automation, Tsinghua University, Beijing 100084, China}
\alsoaffiliation{Institute for Brain and Cognitive Sciences, Tsinghua University, Beijing 100084, China}
\altaffiliation{Shanghai Artificial Intelligence Laboratory, Shanghai 200232, China}
\email{jlsuo@tsinghua.edu.cn}
\author{Qionghai Dai}
\affiliation{Department of Automation, Tsinghua University, Beijing 100084, China}
\alsoaffiliation{Institute for Brain and Cognitive Sciences, Tsinghua University, Beijing 100084, China}

\title[An \textsf{achemso} demo]
  {An Event-Oriented Diffusion-Refinement Method for Sparse Events Completion}


\begin{document}

% \begin{tocentry}

% Some journals require a graphical entry for the Table of Contents.
% This should be laid out ``print ready'' so that the sizing of the

% Inside the \texttt{tocentry} environment, the font used is Helvetica
% 8\,pt, as required by \emph{Journal of the American Chemical

% The surrounding frame is 9\,cm by 3.5\,cm, which is the maximum
% permitted for  \emph{Journal of the American Chemical Society}
% graphical table of content entries. The box will not resize if the
% content is too big: instead it will overflow the edge of the box.

% This box and the associated title will always be printed on a
% separate page at the end of the document.

% \end{tocentry}

\begin{abstract}
  Event cameras or dynamic vision sensors (DVS) record %are bio-inspired sensors with 
asynchronous response to brightness changes instead of conventional intensity frames, and feature ultra-high sensitivity at low bandwidth. The new mechanism demonstrates great advantages in challenging scenarios with fast motion and large dynamic range.  
However, the recorded events might be highly sparse due to either limited hardware bandwidth or extreme photon starvation in harsh environments. 
To unlock the full potential of event cameras, we propose an inventive event sequence completion approach conforming to the unique characteristics of event data in both the processing stage and the output form. 
Specifically, we treat event streams as 3D event clouds in the spatiotemporal domain, develop a diffusion-based generative model to generate dense clouds in a coarse-to-fine manner, and recover exact timestamps to maintain the temporal resolution of raw data successfully. 
To validate the effectiveness of our method comprehensively, we perform extensive experiments on three widely used public datasets with different spatial resolutions, and additionally collect a novel event dataset covering diverse scenarios with highly dynamic motions and under harsh illumination. Besides generating high-quality dense events, our method can benefit downstream applications such as object classification and intensity frame reconstruction.
% In the future, we expect event data analysis 
% the development of point cloud algorithms can benefit the analysis of event data in more aspects, such as classification, segmentation and object recognition.

% We also show our method can benefit downstream applications such as frame reconstruction and object detection. By relating event sequences to point clouds, we expect the development of point cloud algorithms can benefit the analysis of event data in more aspects, such as classification, segmentation and object recognition. Code will be released on XXX.
% To explore the real application of event cameras, we captured a novel event dataset with ground truth and various challenge levels. Extensive experiments on it and a variety of other event datasets with different resolutions demonstrate the effectiveness and applicability of our method.

\end{abstract}

\section{Introduction}
As a novel bio-inspired sensor, event cameras work in a different way from conventional intensity cameras via sensing asynchronous pixel-wise brightness changes. The working principle renders the sensor unique characteristics such as high sensitivity, low latency and high temporal resolution, which provide reliable visual information and wide applications in extreme environments, e.g. fast avoidance~\cite{doi:10.1126/scirobotics.aaz9712}, low-light/high dynamic range perception~\cite{10.1007/978-3-030-58523-5_39} and high-speed imaging~\cite{8946715,Wang_2019_CVPR}~etc. However, such features are traded with spatial and temporal sparsity in the data stream. Firstly, only brightness changes exceeding a threshold can be recorded, and the outputs mostly locate salient moving edges and patterns. Although this issue can be alleviated by raising the sensitivity level but would bring more noise, imposing great challenges to the successive processing and analysis. Secondly, the readout speed may be limited by the hardware's bandwidth (e.g. drone, PC etc.) even if the camera itself works in the full-capacity mode and causes missing entries in the data stream, which is especially severe in high-resolution and busy scenes. The above issues would degenerate or even fail many off-the-shelf event analysis algorithms working well on event streams in ordinary scenarios. To fully utilize the advantages of event cameras in challenging cases (e.g. low-light, high-speed), recovering the missing signals from the sparse recorded event streams is of crucial importance, but remains an under-explored area.

\begin{figure}[]
\vspace{-2mm}
    \centering
    \includegraphics[width=0.56\textwidth]{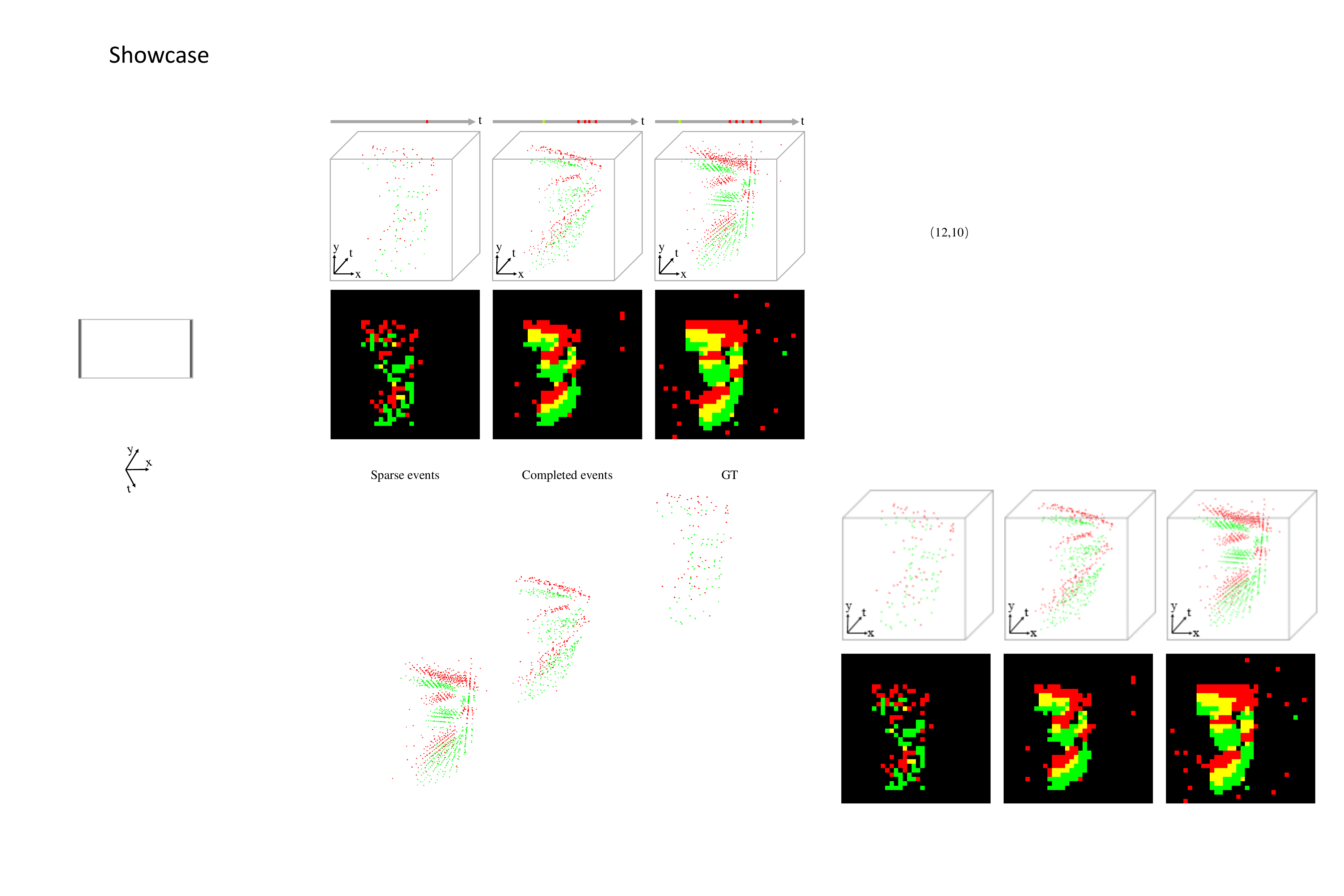}
    % \vspace{-1mm}
    \caption{An exemplar demonstration of our event completion performance, in terms of 3D spatiotemporal cloud (upper) and accumulated 2D image (lower). left: the sub-sampled sparse sequence consisting of 128 events; middle: the completed counterpart; right: the ground truth.}
    \label{fig:showcase}
    \vspace{-3mm}
\end{figure}

In analog to other event quality enhancement tasks, such as super-resolution ~\cite{Duan_2021_CVPR,LI2019206,Li_2021_ICCV,Wang_2020_CVPR}, joint denoising and super-resolution ~\cite{Wang_2020_CVPR}~\cite{Duan_2021_CVPR}, one can convert raw events into 2D grid-based representation for algorithm development. This intuitive solution facilitates adapting the algorithms working on conventional image/video frames, but faces limitations in multiple aspects: assigning no or random timestamps to the output events would lose the temporal ordering information; the output event frames are non-binary, which deviates from the format of event data; the grid-based representation includes large proportion of event-free elements and thus the successive algorithms are storage demanding; such mismatch between the representation and the intrinsic structure might further lead to artificial results in recovered event streams and even harm the downstream analysis.
%Since the low spatial resolution of event cameras leads to degraded event signals, an intuitive practice to enhance the quality of event streams is to super-resolve the raw data. There are a few works that propose to raise the spatial resolution of low-resolution input events~\cite{Wang_2020_CVPR,Duan_2021_CVPR,LI2019206,Li_2021_ICCV}. Wang~et al.~\cite{Wang_2020_CVPR} and Duan~et al.~\cite{Duan_2021_CVPR} proposed methods to jointly denoise and super-resolve event sequences via a guided-filtering and a deep learning framework respectively. However, they both convert raw events into grid-based frames and assigning no or random timestamps to the output super-resolved events. This practice would cause temporal information to be lost and the output event frames are non-binary, deviating from the original event data format. In addition to more storage occupation, such practice may lead to artificial results that make downstream analysis and applying off-the-shelf event algorithms indirect. 
In comparison, Li~et al.~\cite{Li_2021_ICCV} proposed an inspiring strategy to super-resolve the events while maintaining the temporal information. However, the temporal precision is limited to milliseconds when using spiking neural network (SNN) for simulation, and far insufficient for microsecond responses of event cameras. An efficient algorithm making extensive use of the unique structure of event sequence and conforming to its format is highly demanded. 

Event sequence can be formulated as a binary 3D data given a time duration (shown in the left of Fig.~\ref{fig:showcase}), with a four-element tuple $(x, y, t, p)$ denoting the location, time instant and polarity of each event. %Specifically, each event can be seen as a point with binary value in the 3D coordinate system $(x, y, t)$ and a event sequence 
In other words, the event occurrences compose a cloud, similar to the point cloud in 3D vision. %In this paper, we propose to generate a complete representation of the observation from the given sparse observation of events. 
Built on this representation, we propose an event data completion method based on the powerful generative discrete diffusion probabilistic model (DDPM), and develop an event-oriented deep network as the cornerstone. Our method works in a coarse-to-fine manner that firstly predicts a coarse distribution on the condition of sparse event sequences and then refines the generated events with the conditional input with a second sub-network. The final output of the network is a completed set of 3D events that can be transformed back to sequential format without losing temporal ordering information (middle column, Fig.~\ref{fig:showcase}). To validate our effectiveness in diverse scenarios, we collect a new dataset consisting of diverse challenging scenes and conduct experiments on it together with three public datasets with different spatial resolutions. Furthermore, we also show that our method can benefit downstream applications, including object classification and frame reconstruction. 

In sum, this paper contributes in the following aspects:
\begin{itemize}
\vspace{-3mm}
\setlength{\parskip}{0pt}
\setlength{\parsep}{0pt}
\setlength{\itemsep}{0pt}
    \item We propose to represent an event stream as a cloud and recover the dense event signals underlying the recorded sparse event streams via a diffusion-based generative model. 
    \item We develop an event-oriented network as the cornerstone of the diffusion model, which outputs complete dense events with better visual quality while maintaining temporal ordering information. 
    \item We validate the advantageous performance of our approach and its wide applicability to diverse scenes on three public datasets with different resolutions, and show our superior performance on real challenging cases with a self-captured dataset.
    \item We conduct two downstream tasks using the completed events, i.e. object classification and intensity frame reconstruction, and obtain satisfying results, demonstrating the wide applications of our method.
\end{itemize}

\begin{figure*}[t]
    \centering
    \includegraphics[width=\textwidth]{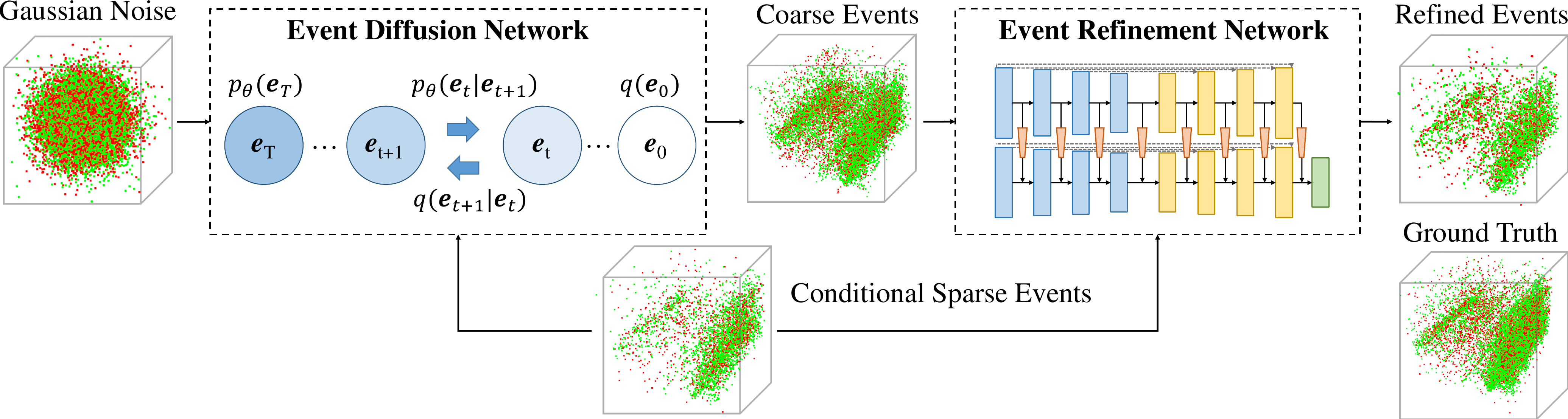}
    % \vspace{-1mm}
    \caption{The overview of diffusion-based coarse-to-fine event completion pipeline. First, we use an event-oriented network to generate coarse distributions of events based on conditional sparse events. Then, we use a second network to yield final completed dense events.}
    \label{fig:overall}
    \vspace{-2mm}
\end{figure*}

\section{Problem Statement}
\noindent\textbf{Event Formation.~~~~}
As an asynchronous sensor, event camera senses pixel-wise illuminance changes, with each pixel independently responding to the change in the logarithmic brightness $L(\bm{q}_k,t_k)\doteq\log(\bm{q}_k,t_k)$. Specifically, at pixel $\bm{q}_k=(x_k,y_k)^T$ and time $t_k$, an event occurs when the brightness change since the last event at this location reaches a threshold $\pm T (T>0)$
\begin{equation}\label{eq:formation}
\textstyle L(\bm{q}_k,t_k)-L(\bm{q}_k,t_k-\Delta t_k)\geq p_kT,
\end{equation}
where $\Delta t_k$ is the time since the last event at $\bm{q}$ and $p_k$ is a binary value (1 or -1) indicating the sign of the change in brightness. An event sequence can be represented as a set of four-element tuples $\varepsilon(t_N)=\lbrace e_k\rbrace_{k=1}^N=\lbrace(t_k,x_k,y_k,p_k)\rbrace_{k=1}^N$ with microsecond resolution.

\noindent\textbf{Event Completion Formulation.~~~~}
Sparsity is the intrinsic characteristic of event data, while too sparse events contain limited information for any application. The event completion task arises when event cameras capture insufficient events in challenging environments such as high-speed and dark scenarios, especially for a large-pixel-number sensor which would also encounter extreme spatial sparsity.
% For example, an event frame captured by a DVXplorer suffer from loss of events due to limited bandwidth in high-speed and low-light scenarios, as shown in Fig.~\ref{}.
In Eq.~\ref{eq:formation}, the threshold $T$ is corresponding to the sensitivity of the sensor. Physically raising $T$ can raise the density of the sensed events but also induces more noise, which is a fundamental trade-off for event camera.

For this task, suppose $\bm{e}_L$ and $\bm{e}_H$ denote the events captured with $T_L$ and $T_H$ for the same scene, where $T_L<T_H$ from the latent clean dense events $\bm{e}_{CD}$
\begin{equation}
\bm{e}_L=\mathcal{C}(\bm{e}_{CD},T_L,\delta)\ \textstyle or\ \bm{e}_H=\mathcal{C}(\bm{e}_{CD},T_H,\delta),
\end{equation}
% \begin{equation}
% \bm{e}_H=\mathcal{C}(\bm{e}_{CD},T_H,\delta),
% \end{equation}
where $\mathcal{C}$ is the capture process of the sensor and $\delta$ denotes the camera settings except sensitivity.
As analyzed above, $\bm{e}_L$ contains clean sparse events while $\bm{e}_H$ contains noisy dense events. The objective of enhancing event quality is to recover clean dense events $\bm{e}_{CD}$ from either of the captured degraded inputs, i.e. $\bm{e}_L$ or $\bm{e}_H$. Event denoising task is defined as recovering $\bm{e}_{CD}$ from $\bm{e}_H$.
Similarly, event completion task can be defined as the recovery of $\bm{e}_{CD}$ from the clean sparse observation $\bm{e}_L$
\begin{equation}\label{eq:formulation}
\hat{\bm{e}_{CD}}=\mathcal{F}(\bm{e}_{L},\theta),
\end{equation}
where $\mathcal{F}$ denotes the reconstruction algorithm and $\theta$ represents its parameters.
% and $T_H$ is the optimal sensitivity for the scene, which can capture clean and dense events. 
% The task can be formulated as the recovery of $\bm{e}_H$ from the observation $\bm{e}_L$.
For most of the time, the paired $\bm{e}_{CD}$ and $\bm{e}_L$ cannot be acquired simultaneously, so we use a random sampling strategy to simulate sparse events from real dense events. Given a complete event set $\bm{e}_{CD}$, the task is to reconstruct $\bm{e}_{CD}$ from the down-sampled event set $\mathcal{S}(\bm{e}_{CD})$. Therefore, Eq.~\ref{eq:formulation} turns into
\begin{equation}
\hat{\bm{e}_{CD}}=\mathcal{F}(\mathcal{S}(\bm{e}_{CD}),\theta).
\end{equation}
% $\textstyle L(\bm{q}_k,t_k)-L(\bm{q}_k,t_k-\Delta t_k)\geq p_kT$

\section{Related work}
\subsection{Event Representation and Quality Enhancement}
\noindent\textbf{Event Representation.~~~~}
% Distinct from intensity cameras, event cameras convert pixelwise logarithmic brightness changes  into the output signal. For a location in the pixel array, 
Event signals have been proven to provide auxiliary help in video deblurring and frame interpolation~\cite{9962797,10.1007/978-3-030-58598-3_41,Tulyakov_2021_CVPR}, image reconstruction and super-resolution~\cite{I._2020_CVPR,Han_2021_ICCV,Wang_2020_CVPR}, and downstream applications such as object recognition~\cite{LiYi_2021_ICCV,Kim_2021_ICCV} and detection~\cite{BMVC2017_40,Cannici_2019_CVPR_Workshops}. With the rapid development of deep learning, many network architectures %have been proposed for pattern recognition or image restoration. To leverage existing state-of-the-art networks, a number of methods 
that embed event streams for either image restoration or pattern recognition have been proposed, such as HFirst~\cite{7010933}, event frame~\cite{Rebecq_2017_BMVC}, event histograms~\cite{Sironi_2018_CVPR}, event-based time surfaces~\cite{7508476}, event spike tensor~\cite{Gehrig_2019_ICCV} and event volume~\cite{Zhu_2019_CVPR}~etc. Among these methods, temporal ordering plays an important role in the effective representation and can influence the performance of downstream applications~\cite{9749022,NEURIPS2020_c2138774,Zhu_2018}.

\vspace{1mm}
\noindent\textbf{Event Quality Enhancement.~~~~}
Raw event signals suffer from severe noise and spatio-temporal sparsity, which challenges the visualization, analysis and downstream applications. If the camera operates in extreme cases, the quality will dramatically decrease further. To address the heavy noise, a number of methods have been proposed to denoise raw event sequences~\cite{8244294,Baldwin_2020_CVPR,9091226,9893571}. Other researchers attempt to super-resolve the raw events by enhancing the spatial resolution~\cite{Wang_2020_CVPR,Duan_2021_CVPR,LI2019206,Li_2021_ICCV}. 
%Wang et al.~\cite{Wang_2020_CVPR} proposed a guided event filter framework that enables high-resolution noise-robust imaging. 
Considering the noise would hamper super-resolution, Duan~et al.~\cite{Duan_2021_CVPR} proposed a deep-learning method to jointly denoise and super-resolve neuromorphic events using an encoder-decoder network, which takes the temporally binned events as input and allocate random timestamps to the output high-resolution events. Such irreversible practice will lose temporal ordering information in the output and may harm downstream applications. As the first attempt to super-resolve events while keeping timestamps, Li~et al.~\cite{LI2019206} proposed a two-stage scheme that first acquires spatial event-count map %with a sparse representation based method 
and temporal rate function,% via a spatio-temporal filter
and then obtains the event of each new pixel with a thinning based event sampling algorithm. 
Further, Li~et al.~\cite{Li_2021_ICCV} proposed a spatio-temporal constraint learning method that optimizes the spatial and temporal event distribution based on SNN model and a simple three-layer CNN. This method achieves pleasant visual quality but requires sufficient events in the sparse input to learn the spatio-temporal distribution and is limited in millisecond resolution due to the numerical simulation of SNN~\cite{NEURIPS2018_82f2b308}. Therefore, an event-to-event recovery method maintaining spatial distribution and sharp details, high temporal resolution and ordering information is highly desired.% for applying event cameras to real challenging scenarios.

\subsection{Point Cloud Completion}
With the maturity of 3D sensors, point clouds have become an important form of modeling 3D scenes. %Relevant approaches have been developed for point cloud analysis, such as classification and segmentation~\cite{Qi_2017_CVPR,NIPS2017_d8bf84be,10.1145/3326362,Zhao_2021_ICCV}. 
A high quality point cloud is essential for downstream tasks and significant progress has been made in generating a complete point cloud from a degraded input. In the past decades, many algorithms have been proposed by using 3D CNNs~\cite{Dai_2017_CVPR,Han_2017_ICCV}, graph CNNs~\cite{valsesia2018learning,10.1145/3326362}, transformer~\cite{Yu_2021_ICCV}. These methods learn a complete point cloud representation under direct supervision of ground truth data. In a distinct way, generative models~\cite{pmlr-v80-achlioptas18a,Pan_2021_CVPR,Luo_2021_CVPR}~etc. learn a probabilistic distribution as representation. %which is further optimized. 
As a new generative model, the denoising diffusion probabilistic model (DDPM)~\cite{pmlr-v37-sohl-dickstein15,NEURIPS2020_4c5bcfec} decomposes the generation process into multiple steps by learning to steadily denoise the random input noise. Due to its powerful generation capability, diffusion model has been applied for point cloud completion~\cite{Luo_2021_CVPR,Zhou_2021_ICCV,lyu2022a} and achieved the state-of-the-art performance. As found in~\cite{lyu2022a}, a conditional DDPM often generates high-quality complete point clouds %with a good overall distribution 
that uniformly covers the shape of the target object. Inspired by DDPM's advantageous performance and the high similarity between point cloud and event cloud, we introduce a conditional DDPM model with an event-oriented encoder-decoder network to generate a dense event sequence with fine details in a coarse-to-fine manner.

% Recently, the diffusion model gains much attention and achieves state-of-the-art performance in generation associated tasks such as image generation, image inpainting, text-to-image conversion etc. By learning to steadily denoise the random input, diffusion models decompose the generation process into multiple steps. Recently, the diffusion model has been adopted for point cloud completion or generation tasks~\cite{}, providing a novel way.

\begin{figure*}[t]
    \centering
    \vspace{-2mm}
    \includegraphics[width=\textwidth]
    {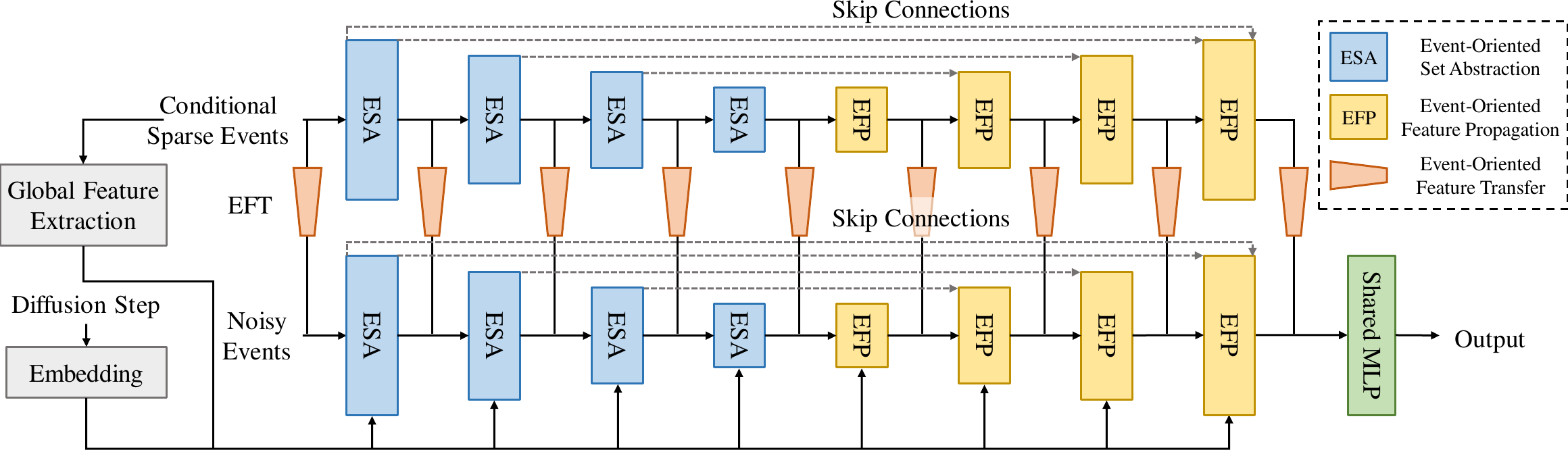}
    % \vspace{-1mm}
    \caption{The architecture of EDR network. The upper branch extracts features from the conditional input, which is absorbed into the lower branch to denoise the noisy input. The proposed event-inspired cuboid query is extensively used in the three main modules---event-oriented set abstraction, feature propagation and feature transfer.}
    \label{fig:network}
    \vspace{-2mm}
\end{figure*}

\section{Method}
In this section, we introduce the event-oriented diffusion refinement (EDR) method, a conditional denoising diffusion probabilistic model for event completion, with the overview illustrated in Fig.~\ref{fig:overall} and the key modules described in the following subsections. %and the formulation, event cloud representation and the proposed deep neural network are described in the following subsections.

\subsection{Event Cloud Representation}\label{sec:event cloud rep}
Raw event data takes the form of a sequence of four-element tuples with each event $\bm{e}_r=(x, y, t, p)$, which are converted into binary points in the 3D coordinate system before being fed into the network for training or inference. Firstly, we cut the event streams sequentially into slices containing $N$ events $\bm{e}_r=\lbrace e_r^i; i=0,\cdots,N\rbrace$, ranked by timestamp. %Since the event streams are ranked by timestamp, each slice consists of $N$ temporally neighboring events. 
Then, the event slice is normalized by the sensor's pixel count along the spatial dimension  and by the time duration along the temporal dimension, i.e.
% i.e., $x^i=(\frac{x_r^i}{W-1}-0.5)\times2, y^i=(\frac{y_r^i}{H-1}-0.5)\times2, t^i=(\frac{t_r^i-t_r^0}{t_r^N-t_r^0}-0.5)\times2, p^i=(p_r^i-0.5)\times2.$
\begin{equation}
x^i=(\frac{x_r^i}{W-1}-0.5)\times2, y^i=(\frac{y_r^i}{H-1}-0.5)\times2, t^i=(\frac{t_r^i-t_r^0}{t_r^N-t_r^0}-0.5)\times2, p^i=(p_r^i-0.5)\times2.
\end{equation}
% \begin{equation}
% t^i=(\frac{t_r^i-t_r^0}{t_r^N-t_r^0}-0.5)\times2
% \label{eq:normalization}
% \end{equation}
So far, we can denote a sample $\bm{e}$ with $N$ events whose $x, y$ and $t$ values are between $-1$ and $1$. These processed events can be regarded as a set of event entries in the 3D coordinate system (similar to point cloud), as shown in Fig.~\ref{fig:showcase}. Besides, the polarity of each point is also attached as its feature. The network completing this set of events is built on this representation and the output conforms exactly to the same form, which can be converted back to the set of four-element tuples ordered by the timestamp.

Despite the high similarity with point cloud, the event cloud %are pre-processed to form points in the 3D coordinate, they 
differs in multiple aspects. First of all, the $t-$ dimension has different metric with spatial dimensions $x-$ and $y-$, thus the event entries are unevenly distributed in the 3D space. Secondly, the normalized event points cannot form a 3D shape with smooth surfaces and have discontinuous and even scattered details instead. Besides, the events has its polarity information which is of specific meanings in physics. Therefore, we need to develop networks matching well with the unique representation.

\subsection{Revisiting Conditional DDPM}\label{sec:ddpm}
The denoising diffusion probabilistic model consists of two processes---diffusion and reverse. In the diffusion process (the blue left-arrow in Fig.~\ref{fig:overall}), Gaussian noise is added to the clean complete events step by step. In the reverse process (the blue right-arrow in Fig.~\ref{fig:overall}), the noise is predicted by the proposed event diffusion network and clean complete events are gradually recovered from the degraded version gradually.
% where a neural network is used to predict the added noise and generate coarse event clouds.

\vspace{1mm}
\noindent\textbf{The diffusion process.~~~~}
Denoting the index of time steps as $t$, the Markov diffusion process from clean complete events $\bm{e}_0$ to $\bm{e}_T$ is defined as
\begin{equation}
q(\bm{e}_1,\cdots,\bm{e}_T)=q(\bm{e}_0)\prod_{t=1}^Tq(\bm{e}_t|\bm{e}_{t-1}),
\label{eq:diffusion}
\end{equation}
where $q(\bm{e}_t|\bm{e}_{t-1})=\mathcal{N}(\bm{e}_t;\sqrt{1-\beta_t}\bm{e}_{t-1},\beta_t\bm{I})$, with the Gaussian noise values $\beta_t$ being pre-defined small positive constants.
Following~\cite{NEURIPS2020_4c5bcfec}, let $\alpha_t=1-\beta_t$ and  $\bar{\alpha}_t=\prod_{i=1}^t\alpha_i$, the diffusion process $q(\bm{e}_t|\bm{e}_0)=\mathcal{N}(\bm{e}_t;\sqrt{\bar{\alpha}_t}\bm{e}_0,(1-\bar{\alpha}_t)\bm{I})$. When $T$ is large enough, $\bar{\alpha}_t$ approaches zero, and $q(\bm{e}_T|\bm{e}_0)$ gets close to the latent distribution which is a Gaussian prior. Then, $\bm{e}_t$ can be sampled with the simplified equation
\begin{equation}
\bm{e}_t=\sqrt{\bar{\alpha}_t}\bm{e}_0+\sqrt{1-\bar{\alpha}_t}\bm{\epsilon}
\label{eq:diffusion2}
\end{equation}
where $\bm{\epsilon}$ is standard Gaussian noise.

% The derivations in~\cite{NEURIPS2020_4c5bcfec} tell when $T$ is large enough, $q(\bm{e}_T|\bm{e}_0)$ approaches the latent distribution which is a Gaussian prior, and then the noisy events $\bm{e}_t$ can be sampled by a defined equation, as detailed in Appendix.

% Following~\cite{NEURIPS2020_4c5bcfec}, let $\alpha_t=1-\beta_t$ and  $\bar{\alpha}_t=\prod_{i=1}^t\alpha_i$, the diffusion process $q(\bm{x}_t|\bm{x}_0)=\mathcal{N}(\bm{x}_t;\sqrt{\bar{\alpha}_t}\bm{x}_0,(1-\bar{\alpha}_t)\bm{I})$. When $T$ is large enough, $\bar{\alpha}_t$ approaches zero, and $q(\bm{x}_T|\bm{x}_0)$ gets close to the latent distribution which is a Gaussian prior. Then, $\bm{x}_t$ can be sampled with the simplified equation
% \begin{equation}
% \bm{x}_t=\sqrt{\bar{\alpha}_t}\bm{x}_0+\sqrt{1-\bar{\alpha}_t}\bm{\epsilon}
% \label{eq:diffusion2}
% \end{equation}
% where $\bm{\epsilon}$ is standard Gaussian noise.

\vspace{1mm}
\noindent\textbf{The reverse process.~~~~}
The reverse process is also a Markov process in which the added noise is predicted and removed afterwards. Conditioned on the input sparse events $\bm{c}$, the reverse from noisy $\bm{e}_T$ to clean events $\bm{e}_0$ is defined as
\begin{equation}
p_{\bm{\theta}}(\bm{e}_0,\cdots,\bm{e}_{T-1})=p(\bm{e}_T,\bm{c})\prod_{t=1}^Tp_{\bm{\theta}}(\bm{e}_{t-1}|\bm{e}_t,\bm{c}), 
\label{eq:reverse}
\end{equation}
where $p_{\bm{\theta}}(\bm{e}_{t-1}|\bm{e}_t,\bm{c})=\mathcal{N}(\bm{e}_{t-1};\bm{\mu}_{\bm{\theta}}(\bm{e}_t,\bm{c},t),\sigma_t^2\bm{I})$, with $\bm{\mu}_{\bm{\theta}}(\bm{e}_t,\bm{c},t)$ and $\sigma_t^2$ denoting the predicted shape from our generative model and the variance, respectively. 
% The denoised events can be generated by first sampling $\bm{e}_T$ from a Gaussian distribution and then progressively sampling $\bm{e}_{t-1}$ from $p_{\bm{\theta}}(\bm{e}_{t-1}|\bm{e}_t,\bm{c})$.
% % for $t=T,\cdots,1$.
To generate a sample conditioned on sparse events $\bm{c}$, we start from sampling $\bm{x}_T$ from a Gaussian distribution and then progressively sample $\bm{x}_{t-1}$ from $p_{\bm{\theta}}(\bm{x}_{t-1}|\bm{x}_t,\bm{c})$ for $t=T,\cdots,1$, and finally obtain $\bm{x}_0$.

\vspace{1mm}
\noindent\textbf{The training process.~~~~}
To simplify the training objective, we follow Ho~et al.~\cite{NEURIPS2020_4c5bcfec}'s parameterization
$\sigma_t^2=\frac{1-\bar{\alpha}_{t-1}}{1-\bar{\alpha}_t}$ and $\bm{\mu}_{\bm{\theta}}(\bm{x}^t,\bm{c},t)=\frac{1}{\sqrt{\alpha_t}}(\bm{x}_t-\frac{\beta_t}{1-\sqrt{1-\bar{\alpha}_t}}\bm{\epsilon_\theta}(\bm{x}_t,\bm{c},t)$, in which $\bm{\epsilon_\theta}$ is a neural network estimating noise from noisy point cloud $\bm{x}_t$, diffusion step $t$ and the conditioner $\bm{c}$. The objective reduces to
\begin{equation}\mathcal{L}_\text{Diff}(\bm{\theta})=\mathbb{E}_{t\sim \mathcal{U}([T]),\bm{\epsilon}\sim \mathcal{N}(0,1)}\Vert\bm{\epsilon}-\bm{\epsilon_\theta}(\bm{e}_t,\bm{c},t)\Vert^2,
\label{eq:training}
\end{equation}
where $\mathcal{U}([T])$ is the uniform distribution over ${1,\cdots,T}$, $\bm{\epsilon}$ is the added standard Gaussian noise. The neural network $\bm{\epsilon_\theta}$ can be reparameterized to predict the noise added to the clean event set $\bm{e}_0$, which can be used to denoise the noisy event set:
$\bm{e}_t=\sqrt{\bar{\alpha}}\bm{e}_0+\sqrt{1-\bar{\alpha}_t}\bm{\epsilon}$.
During training we use ${l}_2$ loss to penalize the difference between model's output $\bm{\epsilon_\theta}(\bm{e}_t,\bm{c},t)$ and the true noise $\bm{\epsilon}$.

\subsection{Event Diffusion-Refinement Network}
%In this section, we introduce the network used in our event diffusion-refinement (EDR). 
%We use the encoder-decoder backbones widely used for diffusion models. 

\vspace{1mm}
\noindent\textbf{The network design.~~~~}
Considering the high similarity between event cloud and point cloud for shape representation, we make event-oriented adaption to the point-version encoder-decoder network---PointNet++~\cite{NIPS2017_d8bf84be} and use it as the backbone of two sub-networks, i.e. event diffusion network (EDN) and event refinement network (ERN) in Fig.~\ref{fig:overall}, which complete event clouds at coarse and fine scales respectively. The detailed architecture is shown in Fig.~\ref{fig:network}. 
The backbone is composed of three main modules: set abstraction (SA), feature propagation (FP) and feature transfer (FT). 
Specifically, SA module subsamples the input event points and propagates the input features. SA block consists of a grouping layer to query neighbors for each point, a set of shared multi-layer perceptrons (MLPs) to extract features, and a reduction layer to aggregate features within the neighbors. 
FP module consists of a PA-Deconv module to upsample the intermediate event cloud representation, a set of shared MLPs to process features, and an attention mechanism to aggregate features. 
FT module transmits the information from conditional cloud to denoise the noisy input, and also consists of a grouping layer, a shared MLP and an attention mechanism to extract and aggregate features from the condition. 
Besides, we embed the diffusion step in the SA and FP module.

As introduced in Secions~\ref{sec:event cloud rep} and ~\ref{sec:ddpm}, we pre-process a set of events $(x, y, t, p)$ by normalizing the first three elements which fall into the range of -1 to 1 and treating the polarity (-1 or 1) as a feature for each event point. DDPM firstly generates 3D Gaussian noise with a random polarity feature and during model training, the noise is gradually removed and the polarity is predicted as a feature for each generated point. The main structure is the same between EDN and ERN, but the diffusion step is not used in the ERN.

\begin{figure}[t]
    \centering
    \vspace{-3mm}
    \includegraphics[width=0.5\textwidth]{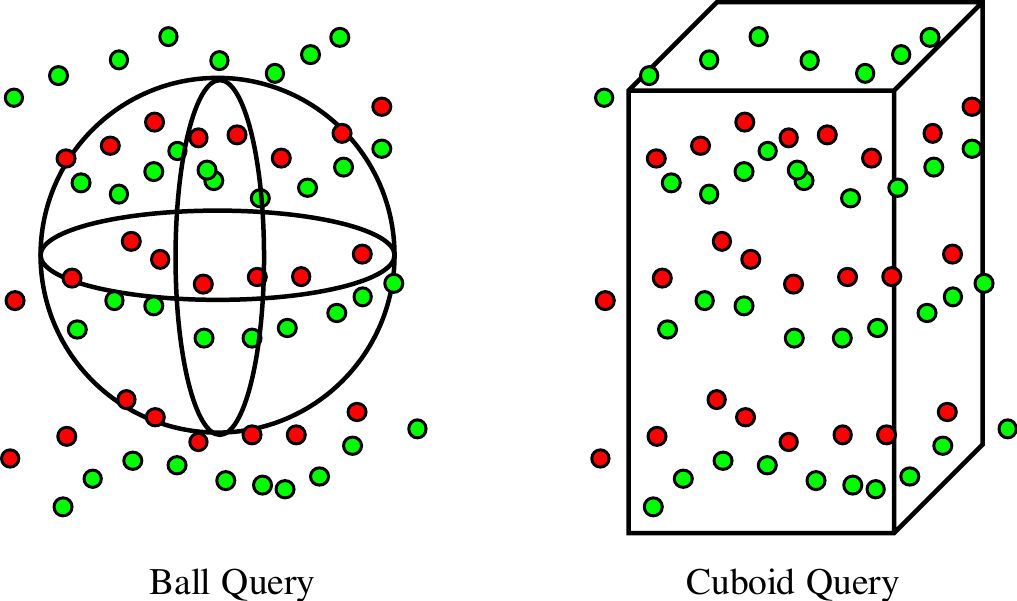}
    \caption{The illustration of the original ball query (left) and the proposed cuboid query (right). Cuboid query consumes more events in the temporal dimension which is important for 3D event cloud representation.}
    \label{fig:cube_query}
    \vspace{-2mm}
\end{figure}

%Therefore, we make the following improvements and form an event-oriented structure.
To match the metric difference between spatial and temporal dimension of the event cloud, we first propose to use a cube query instead of the ball query or KNN query, encouraging the network to aggregate the events in a cube rather than a ball. In this way, the aggregated events resemble the overall distribution of all events and the network is expected to learn a better representation.
Further, we lengthen the cube query along $t-$ dimension, as shown in Fig.~\ref{fig:cube_query}, to let the network pay more attention to the temporally neighboring events than those along $x-$ and $y-$ dimensions, because temporally adjacent events are more informative for event completion.
% (More results in SM) 
% Secondly, we use normalization for receptive field scaling. Since the relative coordinates makes the network optimization harder~\cite{qian2022pointnext}, we use a relative position normalization to make our network adapt to event data with various resolutions. Thirdly, the original PointNet++ is relatively small in capacity. We further add Inverted Residual MLP (InvResMLP) blocks~\cite{qian2022pointnext} to increase the depth of the network, making it to learn fine details in high resolution and large point number scenarios.
%Our proposed improvements are demonstrated to generate complete events with better distribution and quality.

\vspace{1mm}
\noindent\textbf{The network learning.~~~~}
We use the proposed EDN to generate coarse complete events and ERN for refinement. The latter predicts the relative displacement and add it to the coarse events to obtain the refined version. We use the Chamfer Distance (CD) loss between the refined event set $\bm{x}$ and ground truth $\bm{e}$ to supervise the learning of ERN $\bm{\epsilon_f}$
\begin{equation}
\mathcal{L}_{\text{CD}}(\bm{x},\bm{e})\!=\!\frac{1}{|\bm{x}|}\sum_{x\in\bm{x}}\min_{e\in\bm{e}}\Vert x-e\Vert^2\!+\!\frac{1}{|\bm{e}|}\sum_{e\in\bm{e}}\min_{x\in\bm{x}}\Vert x-e\Vert^2,
\label{eq:cd_loss}
\end{equation}
where $|\bm{x}|$ denotes the number of events in $\bm{x}$. As the generation process is slow, we adopt a fast sampling algorithm~\cite{lu2022dpmsolver} to generate and save the coarse events in advance. This practice endures small performance drop compared with 1000-step generation but offers a 99.7\% speedup.

\section{Experiments}
\subsection{Datasets}
To quantitatively evaluate our method and baseline methods, we perform extensive experiments on three public event datasets, i.e. N-MNIST~\cite{10.3389/fnins.2015.00437}, Event Camera Dataset~\cite{Mueggler:228465}, 1Mpx Detection Dataset~\cite{NEURIPS2020_c2138774} at different spatial resolutions. We also collect a dataset to test the performance in diverse real challenging scenarios.
% These data are in variant spatial resolutions---$34\times 34$, $180\times 240$, $720\times 1280$ and $480\times 640$ respectively.

\vspace{1mm}
\noindent\textbf{N-MNIST.~~~~} N-MNIST is an event version of MNIST dataset, which contains around 50000 training samples and 10000 test samples with 10 classes of digits, and the spatial resolution is $34\times 34$. We use 1024 events as the ground-truth and 256/128 events as incomplete input.

\vspace{1mm}
\noindent\textbf{Event Camera Dataset.~~~~} Event Camera Dataset is composed of events captured in daily scenarios with $180\times 240$ resolution. To avoid repetitive scenes, we select 11 snippets (50552 samples) for training and 7 (45388 samples) for test following ~\cite{Li_2021_ICCV}. Since the scenes are of complex structures and with rich semantic information, we use a 50\% sampling rate to down-sample 8192-point ground-truth events to 4096 point sparse input.

\vspace{1mm}
\noindent\textbf{1Mpx Detection Dataset.~~~~} 1Mpx Detection Dataset is captured in a driving environment with a $720\times 1280$ spatial resolution sensor and contains complex scenes. Since the original dataset is very large, we use 80000 and 20000 samples for training and test  respectively. The sparse input contains 4096 points and the dense output 16384 points.

\vspace{1mm}
\noindent\textbf{Self-Captured Dataset.~~~~} To evaluate the methods in real challenging scenarios, we capture a new dataset using an iniVation DVXplorer with resolution $480\times 640$, consisting of rich scenes including moving camera, highly dynamic objects, dim illumination etc. %Especially, we capture fast motion scenes by mounting the camera on a fast rotating stage. 
We include data with various challenges for training, and target to recover 16384 events from down-sampled 4096 events during training. The training set contains 21355 sample. We test on a continuous sequence of 4096 events to qualitatively validate the effectiveness of our approach in real scenarios.

\subsection{Baselines and Metrics}
Since there is no published work for event completion to the best of our knowledge, we compare our approach with a couple of closely related methods, including event super-resolution algorithm---STCL~\cite{Li_2021_ICCV} and point cloud completion algorithms---PoinTr~\cite{Yu_2021_ICCV} and VRCNet~\cite{Pan_2021_CVPR}.
STCL is originally proposed for event super-resolution and we modify its last layer to obtain output events with the same resolution as input.
Besides, since STCL only has millisecond resolution, we set the simulation duration as 25ms for 1Mpx Detection Dataset and 50ms for other datasets. 
% During evaluation, we crop the corresponding number of super-resolved events for calculating metrics and comparison.
PoinTr and VRCNet are easier to be adapted for event completion. Considering that they cannot learn the polarity of event points, we assign the polarity for each entry in the completed event set according to its nearest neighbor in the input sparse events.
% They take sparse 3D point clouds as input and output the completed ones with more points. Our method is built on PDR but with advantages of network structure improvement and enhanced sampling speed.

Since CD loss is sensitive to outliers and cannot reflect the overall distribution, we also use Earth Mover Distance (EMD) to evaluate the quality of the completed events. 
EMD loss penalizes the distribution discrepancy between the predicted events $\bm{x}$ and the ground-truth version $\bm{e}$, by optimizing a transportation problem. Specifically, it estimates a bijection $\phi:\bm{x}\leftrightarrow\bm{e}$ between $\bm{x}$ and $\bm{e}$
\begin{equation}
\mathcal{L}_{\text{EMD}}(\bm{x},\bm{e})=\min_{\phi:\bm{x}\leftrightarrow\bm{e}}\sum_{x\in \bm{x}}\Vert x-\phi(x)\Vert_2.
\label{eq:emd_loss}
\end{equation}
% {\color{magenta} need details if use F1 score.}
Comparatively, EMD is more appropriate for measuring the distance between two distributions.

Despite the fact that CD and EMD are originally for measuring point cloud distances and currently, there is no perfect metric for event sequence as far as we know, both of them are able to measure the distance between 3D event data $(x, y, t)$ since raw 4-element tuple are already converted to 3D event points with 1/-1 polarity feature after normalization.
% CD can also penalize the polarity when considering it as an additional data dimension.
By definition, CD loss does not require two event sets to contain the same number of events and can measure data with any dimension number. Therefore, the use of CD for training is feasible in our method. As a supplement, we use EMD to penalize the overall distribution for 3D event points.

\begin{figure}[h]
    \centering
    \subfigure[]{
    \includegraphics[width=0.49\textwidth]
    {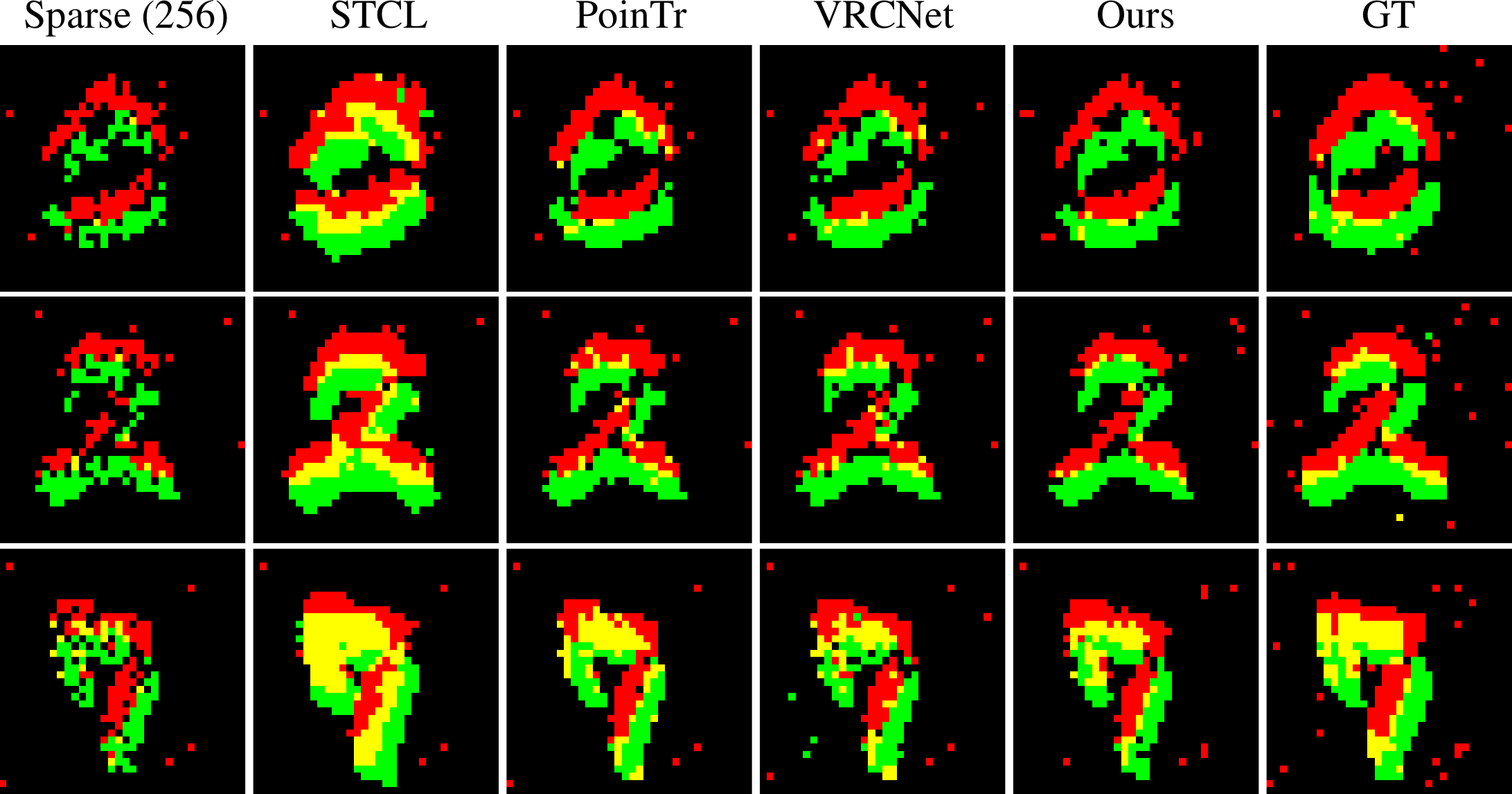}}
    % {figs/crop_res_nmnist_256_nopdr.pdf}}
    \subfigure[]{
    \includegraphics[width=0.49\textwidth]{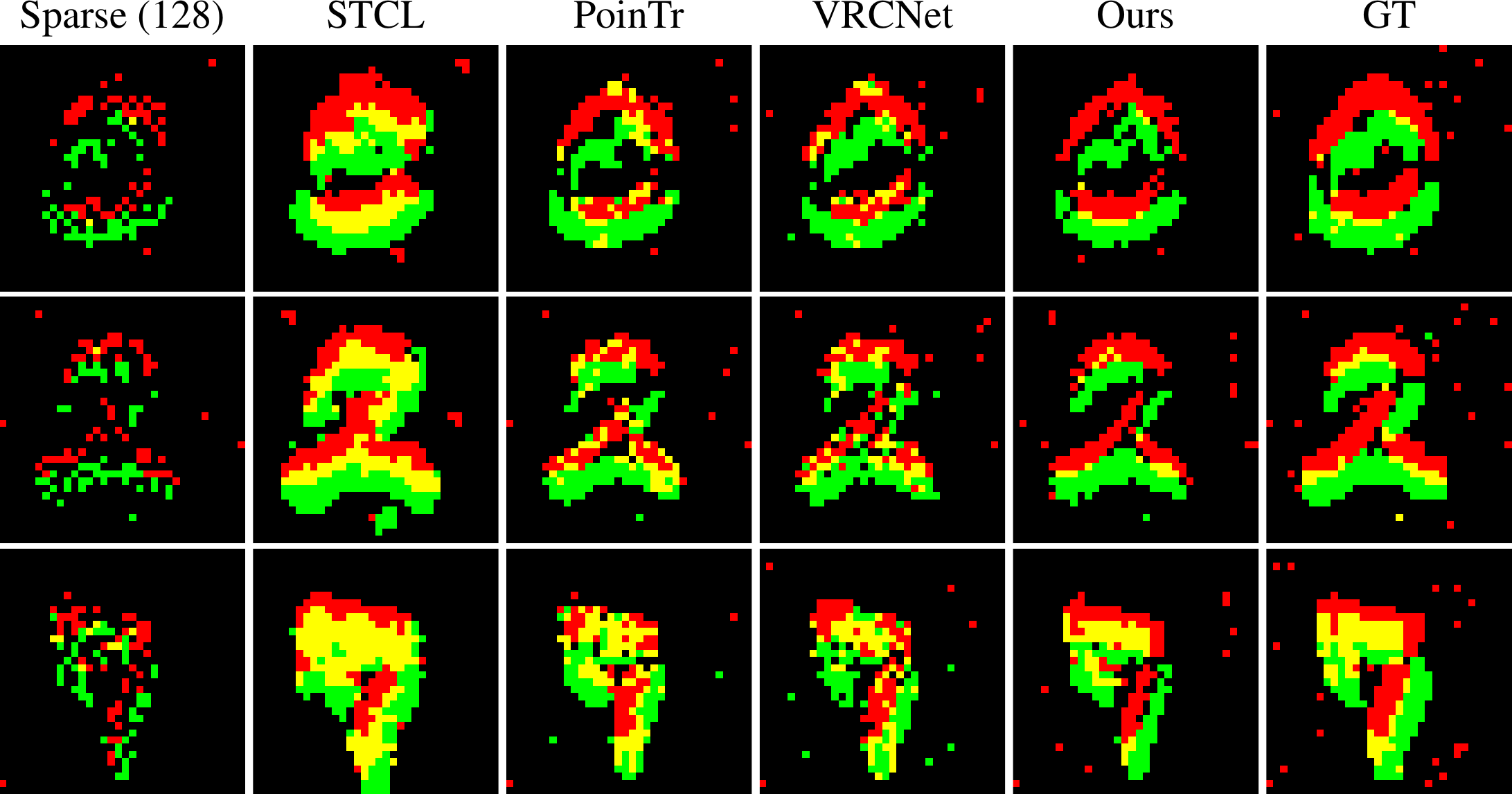}}
    % {figs/crop_res_nmnist_128_nopdr.pdf}}
    % \vspace{-1mm}
    \caption{The event completion results on N-MNIST dataset from input with 256 events (a) and 128 events (b). STCL leads to too dense events which may lose local shape, e.g. '7' in (b), while results of PoinTr and VRCNet tend to suffer from missing entries. Our method maintains both overall event completeness and local shape.}
    \label{fig:res_nmnist}
    %\vspace{-2mm}
\end{figure}

\begin{figure*}[h]
    \centering
    \includegraphics[width=\textwidth]{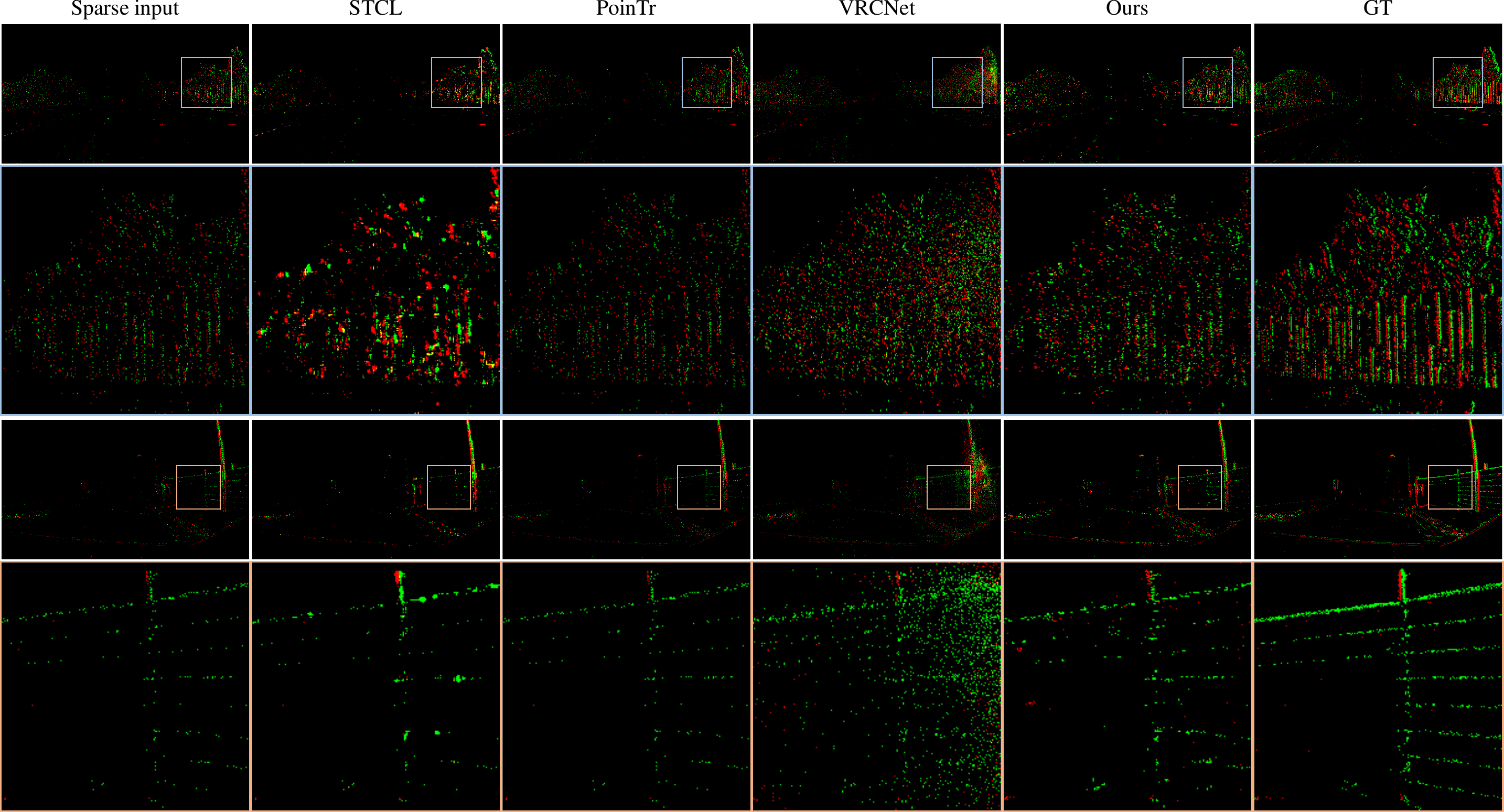}
    % {figs/crop_res_edrive_nopdr.pdf}
    % \vspace{-1mm}
    \caption{The event completion results on examples from 1Mpx Detection Dataset. STCL leads to completed events that tend to gather around certain positions. The results of PoinTr are too sparse, while VRCNet can only learn coarse distribution. In comparison, our method can recover dense events while maintaining sharp details.}
    \label{fig:res_edrive}
    \vspace{-2mm}
\end{figure*}

\begin{figure*}[h]
    \centering    
    \vspace{-3mm}
    \includegraphics[width=\textwidth]{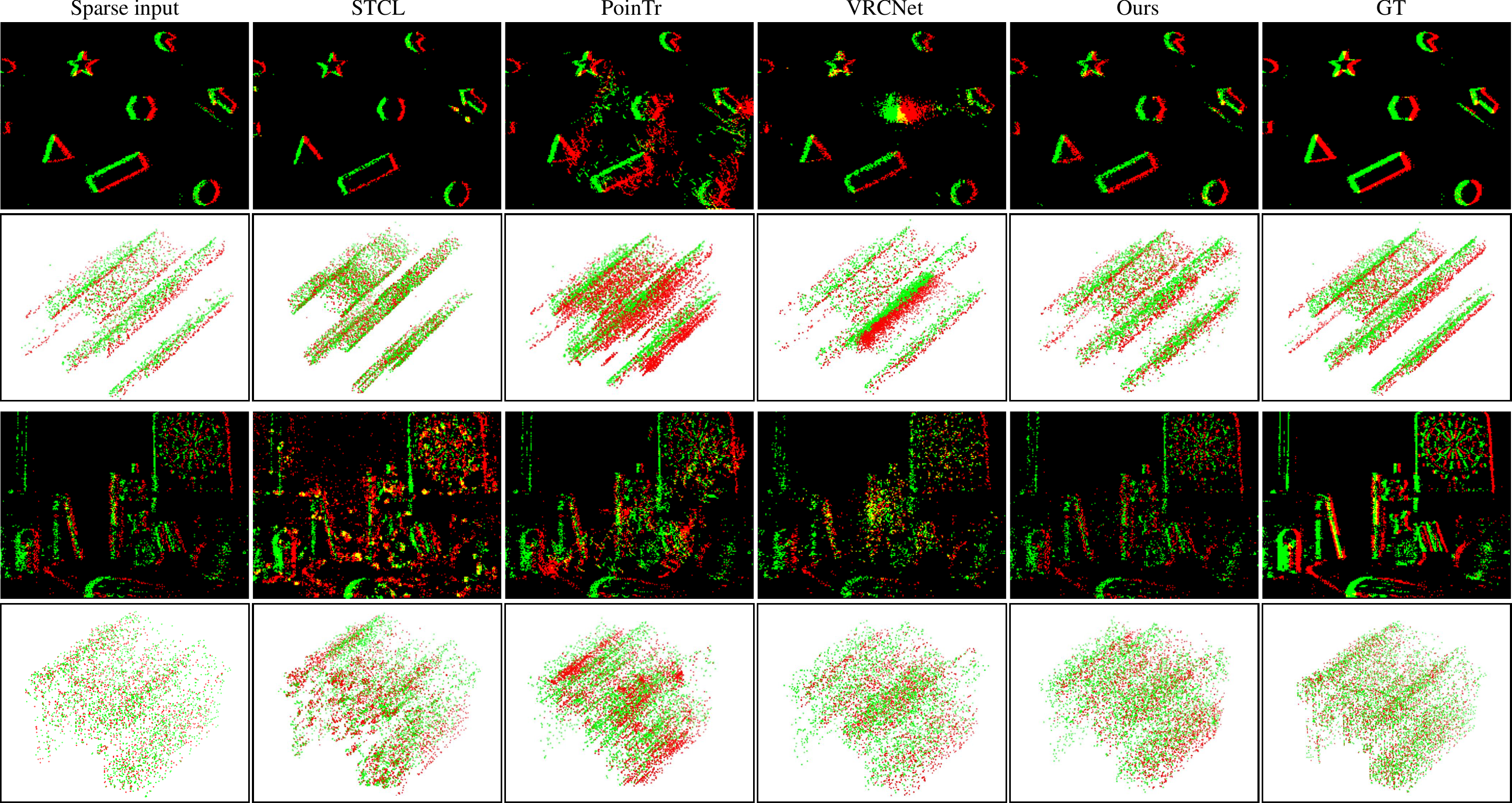}
    % {figs/crop_res_ecd_nopdr.pdf}
    %\vspace{1mm}
    \caption{Visual illustration of event completion results and the reconstructed intensity frames on two examples from Event Camera Dataset. STCL still tends to generate unevenly distributed events gathering together, and STCL and PoinTr are inclined to generate coarse structures and lose details. Our method is free of such artifacts. Intensity frame reconstruction results also validate the superiority of our method.
    }
    \label{fig:res_ecd}
    % \vspace{-1mm}
\end{figure*}

\begin{figure}[h]
    \centering
    \includegraphics[width=0.99\textwidth]
    {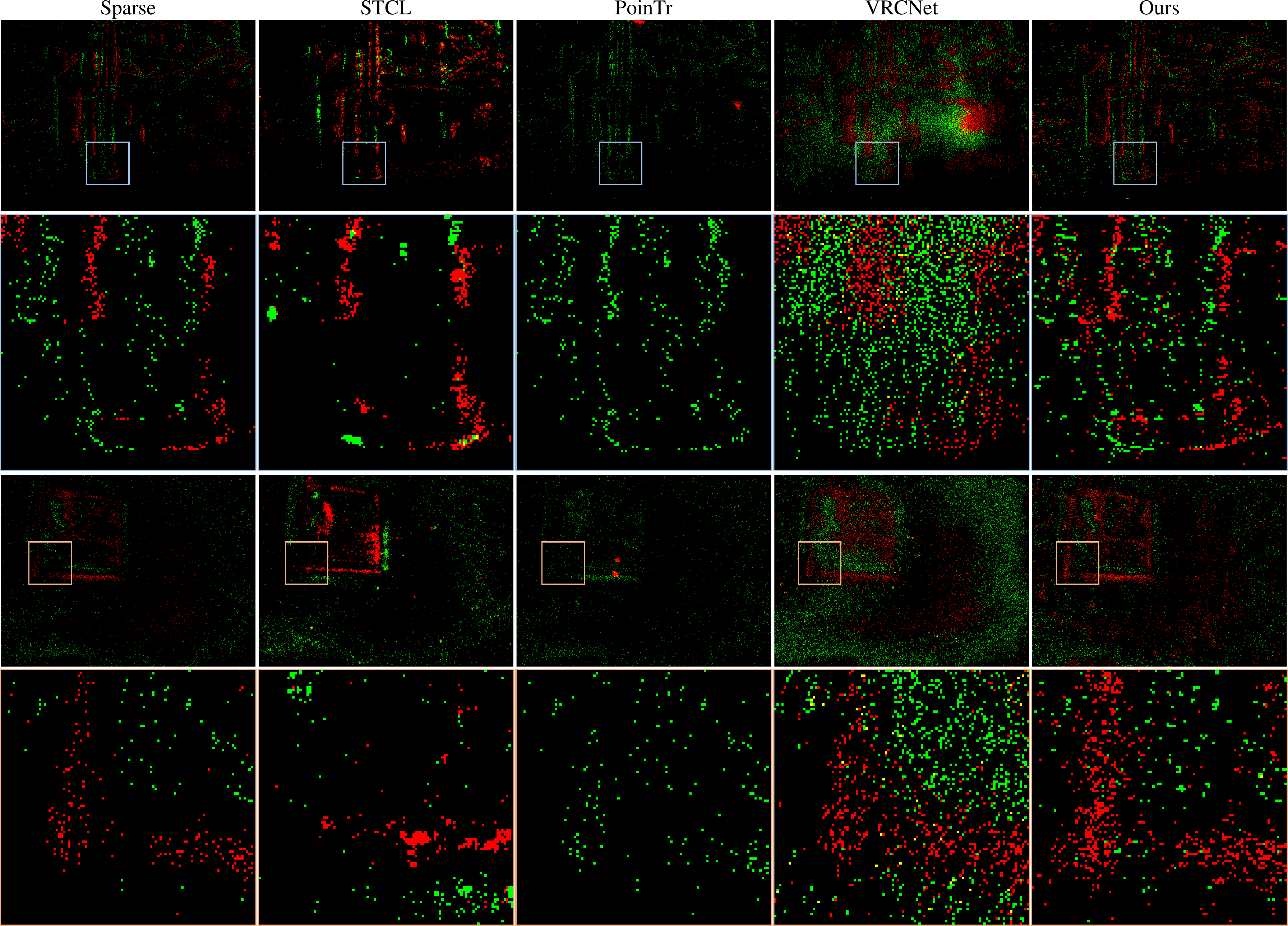}
    % {figs/crop_res_self_vert.pdf}
    % \vspace{-1mm}
    \caption{Visualization of completed events on two examples from the self-captured dataset. left: a tea table captured by a camera on a rotating stage at 1r/s. right: a painting placed in a dark room.}
    \label{fig:res_self}
    %\vspace{-2mm}
\end{figure}

\subsection{Implementation Details}
We learn our model in a coarse-to-fine manner. Firstly, we train the coarse network for 120 epochs for other three public datasets and 300 epochs for the self-captured one, with learning rate of $2e^{-4}$ using Adam optimizer. Since generating a sample with 1000 steps is too time-consuming, we adopt a fast sampling method---DPMSolver~\cite{lu2022dpmsolver} for acceleration and generate a sample after only 27 steps with slight performance degradation. Afterwards, we feed the generated coarse event clouds into the refinement network, which takes 30 epochs to converge.
We empirically found the optimal $t$ of the proposed cuboid query varies for different datasets. Let the bottom edge length be $r$, the optimal length of $t$ dimension is $1r$, $1r$, $1.2r$ and $1.5r$ for the four datasets respectively.

\subsection{Event Completion Results}

% \begin{table*}[t!]
%     \centering
%     \vspace{-3mm}
%     \caption{Comparison of performance. CD loss is multiplied by $10^3$. EMD loss is multiplied by $10^2$.}
%     \vspace{-1mm}
%     \begin{center}
%     % \resizebox{0.98\textwidth}{!}{
%     \begin{tabular}{l|cc|cc|cc|cc|cc}
%         \midrule
%         %\multirow{2}{*}
%          & \multicolumn{2}{c|}{N-MNIST} & \multicolumn{2}{c|}{N-MNIST} & \multicolumn{2}{c|}{Event camera dataset} & \multicolumn{2}{c|}{1Mpx Detection} & \multicolumn{2}{c}{{\color{red}Self-captured}}\\
%         % \toprule
%         & \multicolumn{2}{c|}{1024-256} & \multicolumn{2}{c|}{1024-128} & \multicolumn{2}{c|}{8192-4096} & \multicolumn{2}{c|}{16384-4096} & \multicolumn{2}{c}{16384-4096} \\
%         \cline{2-11}
%         Methods & CD & EMD & CD & EMD & CD & EMD & CD & EMD & CD & EMD \\
%         \toprule
%         %\midrule
%         STCL~\cite{Li_2021_ICCV} & 14.06 & 17.34 & 18.05 & 18.20 & 55.71 & 27.19 & 13.71 & 24.09 & 11.87 & 3.86 \\
%         PoinTr~\cite{Yu_2021_ICCV} & \bf{7.60} & 14.08 & \bf{8.44} & 21.57 & 5.76 & 46.86 & 5.37 & 150.04 & 3.69 & 61.33 \\
%         \midrule
%         Ours (ball) & 7.84 & 7.65 & 9.32 & 10.29 & 3.16 & 30.19 & 3.39 & 32.32 & 1.00 & 19.33 \\
%         Ours (cube) & 7.99 & \bf{6.94} & 9.20 & \bf{8.87} & 3.58 & 18.64 &&  \\
%         \midrule
%     \end{tabular}
%     % }
%     \end{center}
%     \label{tab:exp_main}
% \end{table*}

\noindent\textbf{Quantitative Results.~~~~} In Table~\ref{tab:exp_main}, we report the CD and EMD loss of our method and baseline algorithms. %EMD is more appropriate in measuring the distance between two distributions. 
STCL leads to high CD and EMD compared to the point-based method. It is attributed to the fact that STCL is limited to millisecond resolution in SNN simulation, so cannot learn the latent structure of the events sparsely distributed in both spatial and temporal dimensions. %due to very few event points for both each pixel across time and all pixels at one time point when dealing with very sparse event points, it is difficult to . 
Instead, it is more appropriate for recovering high-resolution event points from dense data. 
As modern point completion networks, PoinTr and VRCNet yield low CD loss on the three datasets, since they use CD loss for optimization. However, the EMD losses are very high, which indicates that it encounters difficulty in completing complex event data with high spatial resolution and sharp details.
% The base model obtains low CD loss for all groups of experiments, but leads to relatively high EMD loss for high-resolution data.
%In comparison, the {\color{red}base model} leads to  both low CD and EMD loss across different datasets and resolutions, demonstrating the feasibility of generative model in this challenging task. The probabilistic nature of the generative diffusion model can benefit the prediction of missing events. %Patterns and shapes under the high resolution event sequence can be recovered via the coarse-to-fine training strategy. 
Still, we notice that our event-oriented method achieves the best EMD across all groups of experiments for all datasets and comparative CD loss to the second competitor, which validates the feasibility of the generative model in predicting missing events and demonstrates that event-specific modules can better represent the distribution of event data. In sum, the superiority of our method is attributed to both the generative nature and event-oriented modules.

\vspace{1mm}
\noindent\textbf{Qualitative Results.\quad}
We visualize the completed event data by accumulating the completed events into 2D frames, as shown in Figs.~\ref{fig:res_nmnist},~\ref{fig:res_ecd},~\ref{fig:res_edrive} and~\ref{fig:res_self}. STCL leads to pleasant results for most cases, but it tends to generate occluded structures. In many situations, the predicted events densely gather at certain locations, losing the sharp thin structures. Although PoinTr and VRCNet obtain low CD loss, the visual results are unpleasant. The visualization indicates that PoinTr fails to learn the data shape or distribution for the whole event set, and instead, the adopted point-wise loss misleads the completed points to adhere to the input sparse events. VRCNet learns coarse distribution but fails to recover sharp details. In comparison, our proposed model can recover details and sharp edges for all datasets. Especially on the self-captured dataset, as shown in Fig.~\ref{fig:res_self}, baseline methods fail to complete informative and sharp shapes, but our method achieves promising visual results in challenging high-speed and low-illumination environments. For example, the legs of the tea table, the contents and the frame of the painting are all clearly reconstructed. Based on the platform for data collection, our method supports the completion of event data captured at a rotation speed of $360^\circ$/s and with an illuminance of 3 lux.

\begin{table*}[h]
    \centering
    \vspace{-1mm}
    \caption{Performance comparison between our method and baseline methods. CD loss indicates event-to-event difference and is multiplied by $10^3$. EMD loss penalizes distribution discrepancy and is multiplied by $10^2$. {\bf{Bold}} denotes the best score.}
    \vspace{-1mm}
    \begin{center}
    \resizebox{1\textwidth}{!}{
    \small
    \begin{tabular}{l|cc|cc|cc|cc}
        \midrule
        %\multirow{2}{*}
         & \multicolumn{2}{c|}{~~~~~~~~~~N-MNIST~~~~~~~~~~} & \multicolumn{2}{c|}{~~~~~~~~~~N-MNIST~~~~~~~~~~} & \multicolumn{2}{c|}{~~~~~~Event Camera Dataset~~~~~~} & \multicolumn{2}{c}{~~~~~~1Mpx Detection Dataset~~~~~~} \\
        % \toprule
        ~~~~~~Methods~~~~~~& \multicolumn{2}{c|}{1024-256} & \multicolumn{2}{c|}{1024-128} & \multicolumn{2}{c|}{8192-4096} & \multicolumn{2}{c}{16384-4096} \\
        \cline{2-9}
         & CD$\downarrow$ & EMD$\downarrow$ & CD$\downarrow$ & EMD$\downarrow$ & CD$\downarrow$ & EMD$\downarrow$ & CD$\downarrow$ & EMD$\downarrow$ \\
        \toprule
        %\midrule
        STCL~\cite{Li_2021_ICCV} & 14.06 & 17.34 & 18.05 & 18.20 & 55.71 & 27.19 & 13.71 & 24.09 \\
        PoinTr~\cite{Yu_2021_ICCV} & 7.60 & 14.08 & 8.44 & 21.57 & 5.76 & 46.86 & 5.37 & 150.04 \\
        VRCNet~\cite{Pan_2021_CVPR} & \bf{7.26} & 16.99 & \bf{7.42} & 17.28 & 4.33 & 22.39 & 3.49 & 27.90 \\
        % Base model & 7.84 & 7.65 & 9.32 & 10.29 & \bf{3.16} & 30.19 & 3.39 & 32.32 \\
        Ablation & 7.84 & 7.65 & 9.32 & 10.29 & \textbf{3.16} & 30.19 & 3.39 & 32.32 \\
        \midrule
        Ours & 7.99 & \bf{6.94} & 9.20 & \bf{8.87} & 3.58 & \bf{18.64} & \bf{3.33} & \bf{19.19} \\
        \midrule
    \end{tabular}
    }
    \end{center}
    \label{tab:exp_main}
    \vspace{-3mm}
\end{table*}

\vspace{1mm}
\noindent\textbf{Ablation Study.\quad}
We conduct an ablation study by replacing the event-oriented cuboid query with a ball query. The EMD loss rises by 0.71, 1.42, 11.55, 11.16 on N-MNIST (256), N-MNIST (128), Event Camera Dataset, and 1Mpx Detection Dataset, respectively although CD loss is similar, as shown in Table~\ref{tab:exp_main}. The results quantitatively validate the contribution of event-oriented cuboid query for effective event representation.

\subsection{Downstream Applications}
To further demonstrate the benefits brought by our event completion method with precise timestamps for the downstream tasks, we conduct two downstream experiments---object classification and intensity frame reconstruction, on the completed event streams.

\vspace{1mm}
\noindent\textbf{Object Classification.~~~~} We test the completed results of N-MNIST for digit classification. We train an object classification network for event data~\cite{Gehrig_2019_ICCV} using complete dense events, and test on the generated events by all methods. The results are shown in Table~\ref{tab:exp_recog}. The 1024-event dense cloud achieves 99.1\% accuracy on the test set. For the 256-event setting, three methods achieve over 90\% accuracy except for VRCNet. STCL obtains 96.7\% accuracy, while our model leads to 96.0\% accuracy. When the number of input points decreases to 128, the accuracy of PoinTr declines dramatically, while the other two still result in over 90\% accuracy. The coarse prediction of both shape and polarity induces low classification accuracy in the 128-point case. VRCNet leads to the lowest accuracy in both settings. In comparison, our method can regress the polarity of each generated event point and high-fidelity shape, which assures the accuracy of this task. Compared to STCL, our method is more robust to the sparsity of input events.

\begin{table}[h]
    \centering
    \caption{Comparison of classification accuracy  on the N-MNIST dataset, using the model trained on high-resolution events to test the completed clouds and LR input. {\bf{Bold}} denotes the best score.}
    \vspace{-3mm}
    \small
    \begin{center}
    \begin{tabular}{l|cc}
        \midrule
        & \multicolumn{2}{c}{N-MNIST} \\
        ~~~~~~~~Methods~~~~~~~& ~~~~~~~~1024-256~~~~~~~~ & ~~~~~~~~1024-128~~~~~~~~ \\
        \toprule
        % {\color{red} LR} & 97.1\% & 95.9\% \\
        LR & 53.6\% & 28.6\% \\
        \midrule
        STCL~\cite{Li_2021_ICCV} & \bf{96.7\%} & 90.8\% \\
        PoinTr~\cite{Yu_2021_ICCV} & 92.8\% & 59.4\% \\
        VRCNet~\cite{Pan_2021_CVPR} & 76.2\% & 45.0\% \\
        Ours & 96.0\% & \bf{91.1\%} \\
        % Ours (cube) & 95.0\% & 90.6\% \\
        \midrule
        GT & \multicolumn{2}{c}{99.1\%} \\
        \midrule
    \end{tabular}
    \end{center}
    \label{tab:exp_recog}
    \vspace{-3mm}
\end{table}

\vspace{1mm}
\noindent\textbf{Intensity Frame Reconstruction.~~~~} We reconstruct intensity frames from the completed results of Event Camera Dataset using E2VID~\cite{8946715} and report the PSNR and SSIM~\cite{1284395} of the results from completed events compared with the reference from ground-truth events in Table~\ref{tab:exp_recon}. Our method achieves the best PSNR for most of the settings and the best SSIM in all settings. 
% The visual results are shown in Fig.~\ref{fig:res_ecd}. Compared to baselines, our method leads to fewer artifacts and better visual quality. 

\begin{table}[t]
    \centering
    \vspace{-2mm}
    \caption{Comparison of reconstructions on the Event Camera Dataset in terms of PSNR and SSIM. {\bf{Bold}} denotes the best score.}
    \vspace{-3mm}
    \begin{center}
    % \resizebox{0.98\textwidth}{!}{
    \small
    \begin{tabular}{l|cccc|cccc}
        \midrule
        %\multirow{2}{*}
         & \multicolumn{4}{c|}{PSNR$\uparrow$} & \multicolumn{4}{c}{SSIM$\uparrow$} \\
        % \toprule
        \cline{2-9}
        Scenes & STCL & PoinTr & VRCNet & Ours & STCL & PoinTr & VRCNet & Ours \\
        \toprule
        %\midrule
        \texttt{boxes\_6dof} & 20.50 & 8.91 & 21.39 & \bf{21.69} & 0.33 & \bf{0.46} & 0.36 & \bf{0.46} \\
        \texttt{calibration} & 17.70 & 14.97 & 17.37 & \bf{18.24} & 0.41 & 0.50 & 0.46 & \bf{0.58} \\
        \texttt{dynamic\_6dof} & 16.68 & 16.17 & 17.66 & \bf{18.87} & 0.47 & 0.67 & 0.45 & \bf{0.70} \\
        \texttt{office\_zigzag} & \bf{20.68} & 15.67 & 16.29 & 17.56 & 0.56 & 0.59 & 0.45 & \bf{0.61} \\
        \texttt{poster\_6dof} & 19.47 & 8.81 & 19.36 & \bf{19.87} & 0.37 & 0.52 & 0.39 & \bf{0.54} \\
        \texttt{shapes\_6dof} & \bf{21.11} & 18.57 & 19.58 & 17.42 & 0.80 & 0.82 & 0.83 & \bf{0.88} \\
        \texttt{slider\_depth} & \bf{22.46} & 15.30 & 16.15 & 21.11 & 0.69 & 0.67 & 0.55 & \bf{0.74} \\
         \midrule
        Average & 19.42 & 10.96 & 19.65 & \bf{20.23} & 0.40 & 0.54 & 0.41 & \bf{0.55} \\
        \midrule
    \end{tabular}
    % }
    \end{center}
    \label{tab:exp_recon}
    \vspace{-5mm}
\end{table}

\section{Conclusion}
In this paper, we target for addressing the lacking density of event streams in challenging cases (e.g., high-speed and low-light conditions) by introducing an event-oriented diffusion-refinement method for event completion to rebuild the missing events. We formulate an event stream as a 3D cloud and design an event-oriented conditional diffusion probabilistic model to generate the completed event points in a coarse-to-fine manner. To the best of our knowledge, this is the first work defining and exploring this task. We compare our method with relevant algorithms to validate its superiority both quantitatively and visually. Furthermore, the performance on two downstream applications, i.e. object classification and intensity frame reconstruction, demonstrates the usability of our method. Our approach would unlock the potential of event cameras and broaden their applications. 

Due to the multi-step sampling process during inference, the generation of coarse events is rather slow, so the training/inference of the proposed method cannot be realized on the fly. In the future, faster and better sampling mechanism can be applied to enable end-to-end training/inference, which will further permit real-time event completion such as on-board deployment.

\begin{acknowledgement}
This work is jointly funded by National Natural Science Foundation of China (Grant No. 61931012 and 62088102) and Beijing Natural Science Foundation (Grant No. Z200021).

\end{acknowledgement}

% \begin{suppinfo}

% A listing of the contents of each file supplied as Supporting Information
% should be included. For instructions on what should be included in the
% Supporting Information as well as how to prepare this material for
% publications, refer to the journal's Instructions for Authors.

% The following files are available free of charge.
% \begin{itemize}
%   \item Filename: brief description
%   \item Filename: brief description
% \end{itemize}

% \end{suppinfo}

\bibliography{ref}

\end{document}